\documentclass[10pt,twocolumn,letterpaper]{article}

\usepackage{cvpr}
\usepackage{times}
\usepackage{epsfig}
\usepackage{graphicx}
\usepackage{amsmath}
\usepackage{amssymb}
\usepackage{graphicx}
\usepackage{caption}
\usepackage{balance}

\usepackage{subcaption}

\usepackage[pagebackref=true,breaklinks=true,letterpaper=true,colorlinks,bookmarks=false]{hyperref}

\cvprfinalcopy 

\def\tilde{\raise.17ex\hbox{$\scriptstyle\sim$}}
\ifcvprfinal\fi

\begin{document}
\title{PANDA: Pose Aligned Networks for Deep Attribute Modeling}

\author{
Ning Zhang$^{1,2}$, \;  Manohar Paluri$^1$, \; Marc'Aurelio Ranzato$^1$, \; Trevor Darrell$^2$, \; Lubomir Bourdev$^1$\\
$^1$Facebook AI Research ~~~~~~~~~~
$^2$EECS, UC Berkeley\\
{\tt\small \{nzhang, trevor\}@eecs.berkeley.edu} \:\:\:\:
{\tt\small \{mano, ranzato, lubomir\}@fb.com}
}

\maketitle

\begin{abstract}
We propose a method for inferring human attributes (such as gender,
hair style, clothes style, expression, action) from images of people
under large variation of viewpoint, pose, appearance, articulation and
occlusion. Convolutional Neural Nets (CNN) have been shown to perform
very well on large scale object recognition problems
\cite{krizhevsky}. In the context of attribute classification,
however, the signal is often subtle and it may cover only a small part
of the image, while the image is dominated by the effects of pose and
viewpoint. Discounting for pose variation would require training on
very large labeled datasets which are not presently available.
Part-based models, such as poselets \cite{AttributesPoseletsICCV2011}
and DPM \cite{dpm} have been shown to perform well for this problem
but they are limited by shallow low-level features.  We propose a new
method which combines part-based models and deep learning by training
pose-normalized CNNs. We show substantial improvement
vs. state-of-the-art methods on challenging attribute classification
tasks in unconstrained settings. Experiments confirm that our method
outperforms both the best part-based methods on this problem and
conventional CNNs trained on the full bounding box of the person.
\end{abstract}

\section{Introduction}
\label{sec:intro}
Recognizing human attributes, such as gender, age, hair style,
and clothing style, has many applications, such as facial
verification, visual search and tagging suggestions.  This is,
however, a challenging task when dealing with non-frontal facing
images, low image quality, occlusion, and pose variations.
The signal associated with some attributes is subtle and the image is
dominated by the effects of pose and viewpoint.  For example, consider
the problem of detecting whether a person wears glasses. The signal
(glasses wireframe) is weak at the scale of the full person and the
appearance varies significantly with the head pose, frame design and
occlusion by the hair. Therefore, localizing object parts and
establishing their correspondences with model parts can be key to
accurately predicting the underlying attributes.

Deep learning methods, and in particular convolutional
nets~\cite{Lecun89}, have achieved very good performance on several
tasks, from generic object recognition~\cite{krizhevsky} to pedestrian
detection~\cite{sermanet-cvpr13} and image
denoising~\cite{dnn-denoising}.  Moreover, Donahue \etal \cite{decaf}
show that features extracted from the deep convolutional network
trained on large datasets are generic and can help in other visual recognition problems.  However, as we report below,
they may underperform compared to conventional methods which exploit
explicit pose or part-based normalization.  We conjecture that
available training data, even ImageNet-scale, is presently
insufficient for learning pose normalization in a CNN, and propose a
new class of deep architectures which explicitly incorporate such 
representations.  We combine a part-based representation with
convolutional nets in order to obtain the benefit of both
approaches. By decomposing the input image into parts that are
pose-specific we make the subsequent training of convolutional nets
drastically easier, and therefore, we can learn very powerful
pose-normalized features from relatively small datasets.

Part-based methods have gained significant recent attention as a method to deal
with pose variation and are the state-of-the-art method for attribute
prediction today. For example, spatial pyramid matching
\cite{spm} incorporates geometric correspondence and spatial
correlation for object recognition and scene classification. The DPM model
\cite{dpm} uses a mixture of components with root filter and part
filters capturing viewpoint and pose variations. Zhang \etal proposed
deformable part descriptors \cite{AttributesDPMsICCV2013}, using DPM
part boxes as the building block for pose-normalized representations
for fine-grained categorization
task. Poselets~\cite{BourdevMalikICCV09,  bourdev10} are part
detectors trained on 
positive examples clustered using keypoint annotations; they capture a
salient pattern at a specific viewpoint and pose. Several approaches
\cite{BirdletsFarrellICCV11, PosePoolingKernelsZhangEtalCVPR12} have
used poselets as a part localization scheme for fine-grained
categorization tasks which are related to attribute prediction.
Although part-based methods have been successful on several tasks,
they have been limited by the choice of the low-level features applied to
the image patches.


\begin{figure*}[t]
\centering
\includegraphics[width =0.9\linewidth] {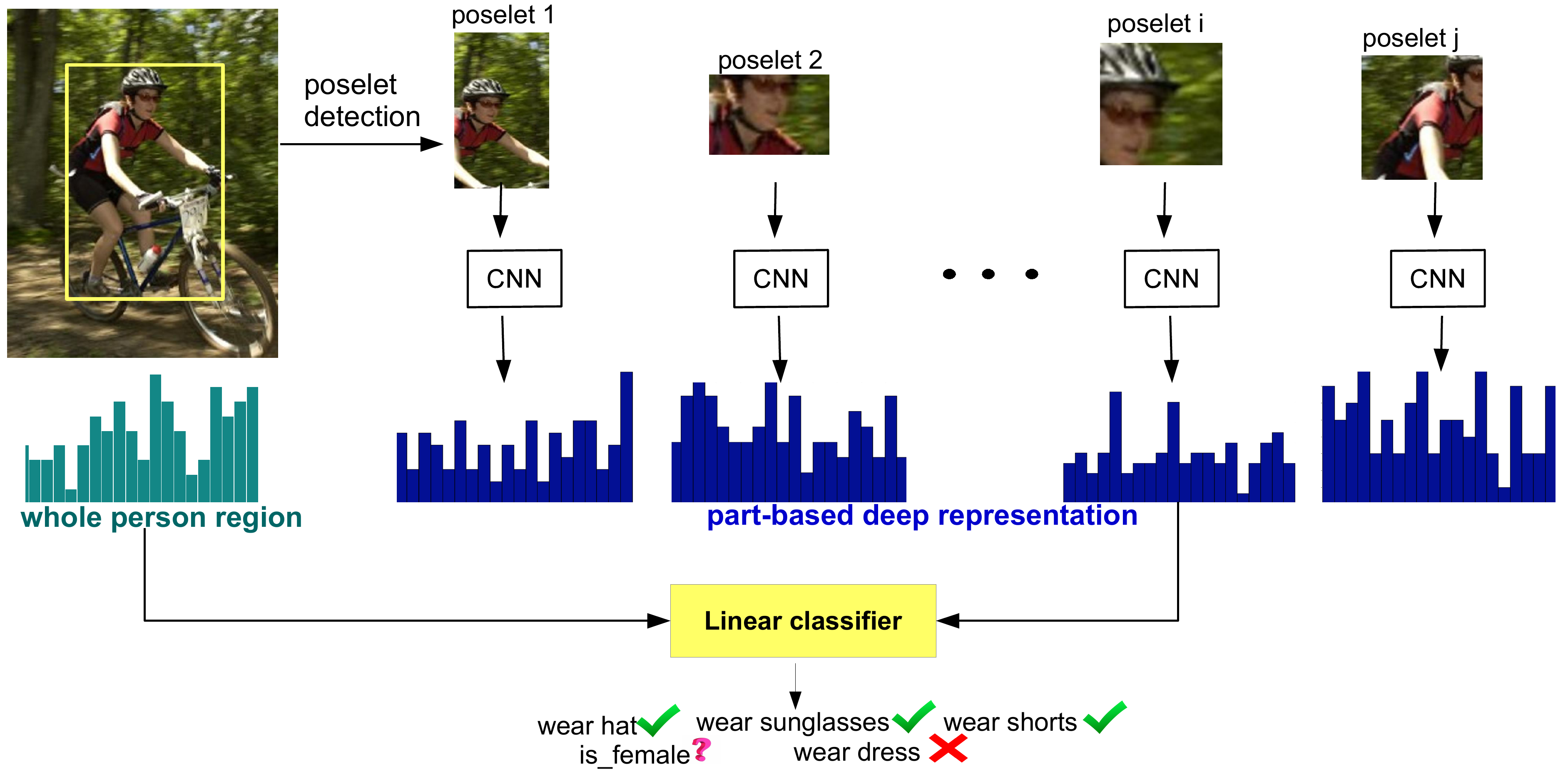}
\caption{\textbf{Overview of Pose Aligned Networks for Deep Attribute modeling (PANDA)}.  One convolutional neural net is trained on semantic part patches for each poselet and then the top-level activations of all nets are concatenated to obtain a pose-normalized deep representation. The final attributes are predicted by linear SVM classifier using the pose-normalized representations.}\label{fig:overview}
\end{figure*}

In this paper, we propose the PANDA model, Pose Alignment Networks for Deep
Attribute modeling, which augments deep convolutional networks to have input layers based on semantically
aligned part patches. Our model learns features that are specific to a
certain part under a certain pose. We then combine the features
produced by many such networks and construct a pose-normalized deep
representation using poselets. Our method can use other parts and we show the performance using DPM~\cite{dpm} as well.  We
demonstrate the effectiveness of PANDA on attribute classification
problems and present state-of-the-art experimental results on four
datasets, an attribute dataset from the web, the Berkeley
Attributes of People Dataset \cite{AttributesPoseletsICCV2011}, the
Labeled Faces in the Wild dataset~\cite{lfw_iccv09}, and a large-scale gender recognition dataset.


\section{Related work}
\label{sec:related}
\subsection{Attribute classification }
Attributes are used as an intermediate representation for knowledge transfer in \cite{lampert_learning_2009, farhadi_describing_2009} for object recognition tasks. By representing the image as a list of human selected attributes they recognize unseen objects with few or zero training examples. Other related work on attributes includes that by Parikh \etal \cite{relative_attributes} exploring the relative strength of attributes by learning a rank function for each attribute, which can be applied to zero-shot learning as well as to generate richer textual descriptions. There is also some related work in automatic attribute discovery: Berg \etal \cite{auto_discovery} proposed automatic attribute vocabularies discovery by mining unlabeled text and image data sampled from the web. Duan \etal \cite{DuanEtalCVPR12} proposed an interactive crowd-sourcing method to discover both localized and discriminative attributes to differentiate bird species. 

In \cite{lfw_iccv09}, facial attributes such as gender, mouth shape, facial expression, are learned for face verification and image search tasks. Some of the attributes used by them are similar to what we evaluate in this work. However, all of their attributes are about human faces and most of images in their dataset are frontal face subjects while our datasets are more challenging in terms of image quality and pose variations. 

A very closely related work on attribute prediction is Bourdev \etal \cite{AttributesPoseletsICCV2011},  which is a three-layer feed forward classification system and the first layer predicts each attribute value for each poselet type. All the predicted scores of first layer are combined as a second layer attribute classifier and the correlations between attributes are leveraged in the third layer.
Our method is also built on poselets, from which the part correspondence is obtained to generate a pose-normalized representation. 

\subsection{Deep learning}
The most popular deep learning method for vision, namely the convolutional neural network (CNN), has been pioneered by LeCun and collaborators \cite{Lecun89} who initially applied it to OCR~\cite{Lecun98OCR} and later to
 generic object recognition tasks~\cite{jarrett-iccv2009}. As more labeled data and computational power has become recently available, convolutional nets have become the most accurate method
 for generic object category classification~\cite{krizhevsky} and pedestrian detection~\cite{sermanet-cvpr13}.

Although very successful when provided very large labeled datasets, convolutional nets usually generalize poorly on smaller datasets because they require the estimation of millions of parameters.
This issue has been addressed by using unsupervised learning methods leveraging large amounts of unlabeled data~\cite{ranzato-cvpr2007, jarrett-iccv2009,googleBrain}. In this work, we take instead a different perspective: we make the learning task easier by providing the network with pose-normalized inputs. 

While there has already been some work on using deep learning methods for attribute prediction~\cite{deep_attribute_networks}, we explore alternative ways to predict attributes, we 
incorporate the use of poselets in the deep learning framework and we perform a more extensive empirical validation which compares against conventional baselines and deep CNNs evaluated on the whole person region.


\begin{figure*}[t]
\centering
\includegraphics[width = \linewidth] {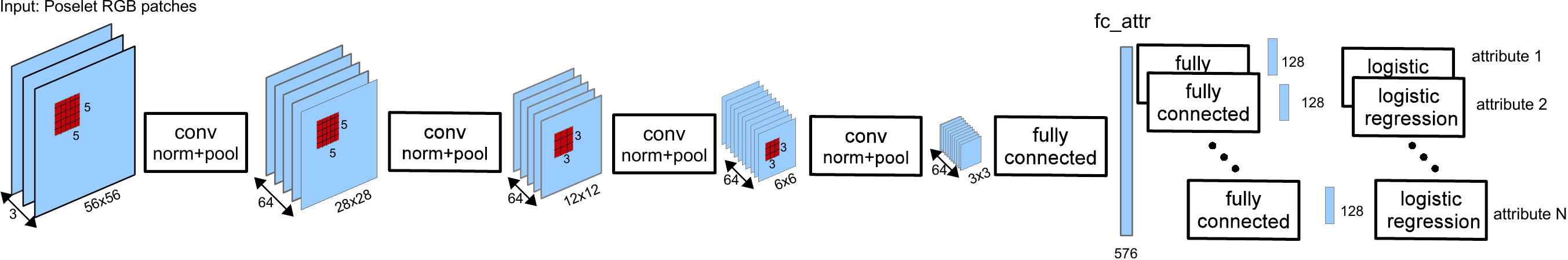}
\caption{\textbf{Part-based Convolutional Neural Nets.} For each poselet, one convolutional neural net is trained on patches resized 64x64.   The network consists of 4 stages of convolution/pooling/normalization and followed by a fully connected layer. Then, it branches out one fully connected layer with 128 hidden units for each attribute. We concatenate the activation from fc\_attr from each poselet network to obtain the pose-normalized representation.
The details of filter size, number of filters we used are depicted above.} \label{fig:outline_hybrid}
\end{figure*}

\section{Pose Aligned Networks for Deep Attribute modeling (PANDA)}
\label{sec:method}
We explore part-based models, specifically poselets, and deep learning, to obtain pose-normalized representations for attribute classification tasks. 
Our goal is to use poselets for part localization and incorporate these normalized parts into deep convolutional nets in order to extract pose-normalized representations. 
Towards this goal, we leverage both the power of convolutional nets for learning discriminative features from data
 and the ability of poselets to simplify the learning task by decomposing the objects into their canonical poses. 
We develop Pose Aligned Networks for Deep Attribute modeling (PANDA), which incorporates part-based and whole-person deep representations.

\begin{figure}[t]
\centering
\includegraphics[width = \linewidth] {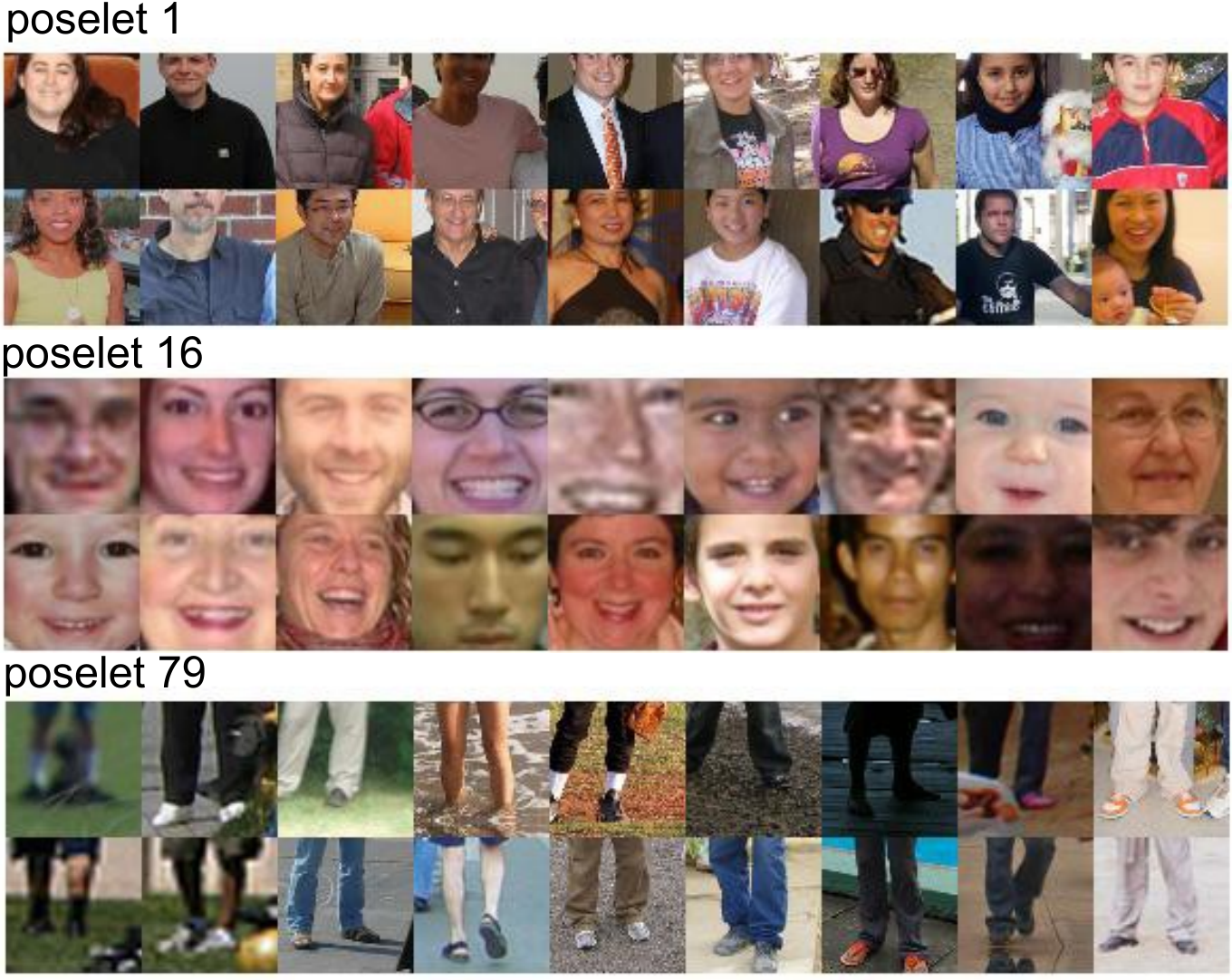}
\caption{\textbf{Poselet Input Patches from Berkeley Attributes of People Dataset.} For each poselet, we use the detected patches to train a convolution neural net. Here are some examples of input poselet patches and we are showing poselet patches with high scores for poselet 1,16 and 79. } \label{fig:poselet_input}
\end{figure}

While convolutional nets have been successfully applied to large scale object recognition tasks, they do not generalize well when trained on small datasets. Our setup requires fewer training instances because we are able to augment the training set size -- we consider each poselet activation as a separate training example.


Specifically, we start from poselet patches, resize them to 64x64 pixels (Figure \ref{fig:poselet_input}), randomly jitter each patch and flip it horizontally with probability 0.5 to improve generalization, and train a CNN for each poselet. The overall convolutional net architecture is shown in Figure \ref{fig:outline_hybrid}. The network consists of four convolutional, max pooling, local response normalization layers followed by a fully connected layer with 576 hidden units.  After that, the network branches out one fully connected layer with 128 hidden units for each  attribute and 
each of the branch outputs a binary classifier of the attribute. The last two layers are split to let the network develop customized features for each attribute (e.g., detecting whether a person
 wears a ``dress'' or ``sunglasses'' presumably  requires different features)
 while the bottom layers are shared to a) reduce the number of parameters and b)
to leverage common low-level structure.  

The whole network is trained jointly by standard back-propagation of the error~\cite{back_propagation}
 and stochastic gradient descent~\cite{SGD} using as a loss function the sum of the log-losses of each attribute for each training sample.
 The details of the layers are given in Figure~\ref{fig:outline_hybrid} and further implementation details can be found in~\cite{krizhevsky}. 
To deal with noise and inaccurate poselet detections, 
we train on patches with high poselet detection scores and then we gradually add more low confidence patches. 

Different parts of the body may have different signals for each of the attributes and sometimes signals coming from one part cannot infer certain attributes accurately. 
For example, deep net trained on person leg patches contains little information about whether the person wears a hat.  
Therefore, we first use deep convolutional nets to generate discriminative image representations for each part separately and then we combine these representations for the final classification. Specifically, we extract the activations from fc\_attr layer in Figure \ref{fig:outline_hybrid}, which is 576 dimensional, for the CNN at each poselet, and concatenate the activations of all poselets together into 576*150 dimensional feature. If a poselet does not activate for the image, we simply leave the feature representation to zero. 

The part-based deep representation mentioned above leverages both the
discriminative deep convonvolutional features and part
correpondence. However, poselet detected parts may not always cover
the whole image region and in some degenerate cases, images may have
few poselets detected. To deal with that, we also incorporate a deep
network covering the whole-person bounding box region as input to our
final pose-normalized representation.

Based on our experiments, we find a more complex net is needed for the
whole-person region than for the part regions.  We extract deep
convolutional features from the model trained on Imagenet
\cite{krizhevsky} using the open source package provided by
\cite{decaf} as our deep representation of the full image patch.

As shown in Figure \ref{fig:overview}, we concatenate the features from the deep representations of the full image patch and the 150 parts and train a linear SVM for each attribute.

\section{Datasets}
\label{sec:dataset}
\subsection{The Berkeley Human Attributes Dataset}

We tested our method on the Berkeley Human Attributes Dataset~\cite{AttributesPoseletsICCV2011}. This dataset consists of 4013 training, and 4022 test images collected from PASCAL and H3D datasets. The dataset is challenging as it includes people with wide variation in pose, viewpoint and occlusion. About 60\% of the photos have both eyes visible, so many existing attributes methods that work on frontal faces will not do well on this dataset.

\subsection{Attributes 25K Dataset}

Unfortunately the training portion of the Berkeley dataset is not large enough for training our deep-net models (they severely overfit when trained just on these images).
We collected an additional dataset from Facebook of 24963 people split into 8737 training, 8737 validation and 7489 test examples.
We made sure the images do not intersect those in the Berkeley dataset. The statistics of the images are similar, with large variation in viewpoint, pose and occlusions.

We train on our large training set and report results on both the corresponding test set and the Berkeley Attributes test set. We chose to use a subset of the categories from the Berkeley dataset and add a few additional categories. This will allow us to explore the transfer-learning ability of our system.

\begin{figure}
\centering
\includegraphics[width=\linewidth]{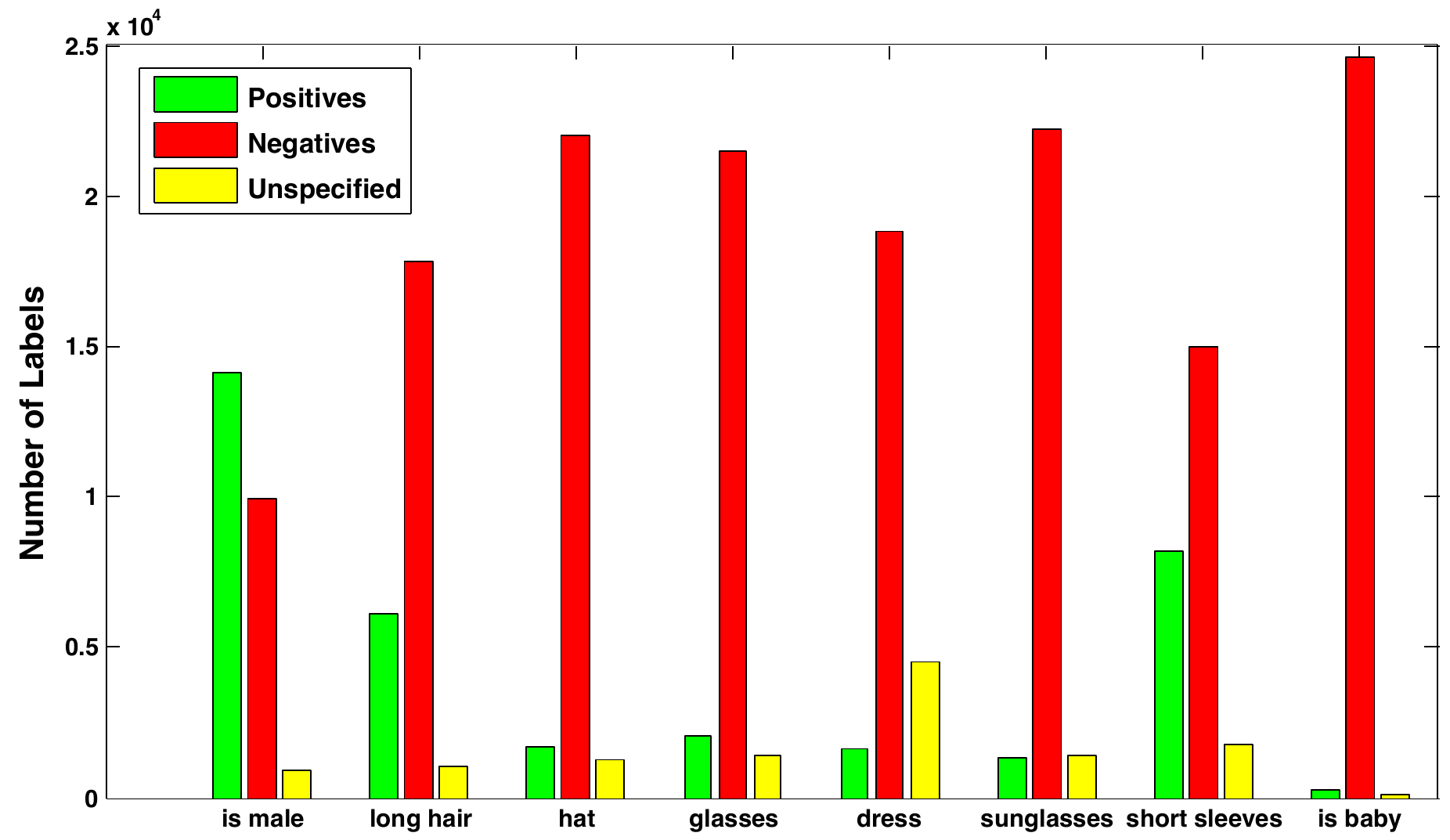}
\caption{Statisitcs of the number of groundtruth labels on Attribute 25k Dataset. For each attribute, green is the number of positive labels, red is the number of negative labels and yellow is the number of uncertain labels.  }\label{fig:fb_statistic}
\end{figure}

Not every attribute can be inferred from every image. For example, if the head of the person is not visible, we cannot enter a label for the "wears hat" category.  The statistics of ground truth labels are shown on Figure~\ref{fig:fb_statistic}.

%
%
%
%
%

\section{Results}

In this section, we will present a comparative performance evaluation of the proposed method. 

\begin{table*}[t]
\centering
\begin{tabular}{|c|c|c|c|c|c|c|c|c|c|c|}
\hline
Attribute	&	male	&long hair	&glasses	&hat 	&tshirt	& longsleeves	&shorts 	& jeans	& long pants & Mean AP \\
\hline
Poselets\cite{AttributesPoseletsICCV2011}&82.4&72.5	&55.6	&	60.1	&	51.2	&74.2	&45.5	&54.7 & 90.3 &65.18\\
\hline
DPD\cite{AttributesDPMsICCV2013}& 83.7	& 70.0	& 38.1	& 73.4	& 49.8	& 78.1	& 64.1	& 78.1 & 93.5 &69.88\\
\hline
Joo \etal \cite{JooICCV2013} & 88.0 & 80.1 & 56.0 & \textbf{75.4} & \textbf{53.5} & 75.2 & 47.6 & 69.3 & 91.1 & 70.7 \\
\hline
PANDA & \textbf{91.7} &  \textbf{82.7} &  \textbf{70.0} &  74.2 & 49.8 &  \textbf{86.0} &  \textbf{79.1} &  \textbf{81.0} &  \textbf{96.4} &  \textbf{78.98} \\
\hline
\end{tabular}
\caption{Attribute classification results on the Berkeley Attributes of People Dataset as compared to the methods of Bourdev~\etal~\cite{AttributesPoseletsICCV2011} and Zhang~\etal~\cite{AttributesDPMsICCV2013} .} \label{table-iccv11}
\end{table*}

\begin{table*}[t]
\centering
\begin{tabular}{|c|c|c|c|c|c|c|c|c|c|}
\hline
Attribute	&	male	&long hair	&hat		&glasses	&dress	&sunglasses	&short sleeves	& baby	& mean AP \\
\hline
Poselets150\cite{AttributesPoseletsICCV2011}&86.00	&75.31	&29.03	&	36.72	&	34.73	&50.16	&55.25	&41.26  &51.06\\
\hline
DPD\cite{AttributesDPMsICCV2013}& 85.84	& 72.40	& 27.55	& 23.94	& 48.55	& 34.36	& 54.75	& 41.38 & 48.60\\
\hline
DeCAF \cite{decaf} & 82.47 &65.03 & 19.15 & 14.91 & 44.68 &26.91 &56.40 & 50.19&44.97 \\
\hline
DL-DPM & 88.27 & 77.64 & \textbf{43.44} & 36.70 & 55.72 & 55.03 & 67.95 & 64.89 & 61.20 \\
\hline
PANDA & \textbf{94.10} 	&\textbf{83.17}	&39.52	&\textbf{72.25}	&\textbf{59.41}	&\textbf{66.62}	& \textbf{72.09}	&\textbf{78.76} &\textbf{70.74} \\
\hline	
\end{tabular}
\caption{Average Precision on the Attributes25K-test dataset.} \label{table-25k}
\end{table*}

\subsection{Results on the Berkeley Attributes of People Dataset}

On Table~\ref{table-iccv11} we show the results on applying our system on the publicly available Berkeley Attributes of People dataset. 
We compare against \textbf{Poselets}, \textbf{DPD} and \textbf{Joo \etal} on that dataset as reported by~\cite{AttributesPoseletsICCV2011}, \cite{AttributesDPMsICCV2013} and \cite{JooICCV2013} respectively. For our method, \textbf{PANDA}, we use Attributes25K train+val dataset to train the poselet-level CNNs of our system, and we used validation set to train the SVM classifier.

As the table shows, our system outperforms all the prior methods across most attributes. In the case of t-shirt,~\cite{JooICCV2013}  performs better, perhaps due to the fact that Attribute25K dataset doesn't have wearing tshirt attribute so that the part-level CNNs are not trained on that attribute. 

Note that the attributes shorts, jeans and long pants are also not present in the Attributes25K dataset. We don't have enough labeled data of this dataset to train the convolutional neural nets but the transfer learning is still effective.

\begin{figure}[t]
\centering
\begin{subfigure}{0.5\textwidth}
\includegraphics[width=\linewidth]{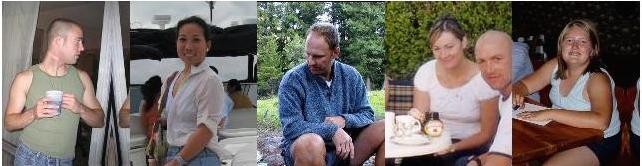}
\caption{Failure case: Top incorrect predictions as wearing tshirts.}
\end{subfigure}
\begin{subfigure}{0.5\textwidth}
\includegraphics[width=\linewidth]{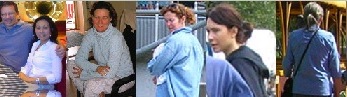}
\caption{Failure case: Top incorrect predictions for short hair.}
\end{subfigure}
\caption{Example of failure cases on the Berkeley Attributes of People test dataset.}\label{fig:failure}
\end{figure}
\begin{figure}[t]
\centering
\begin{subfigure}{0.5\textwidth}
\centering
\includegraphics[width=0.9\linewidth]{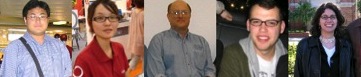}
\caption{Highest scoring results for people wearing glasses.}
\end{subfigure}
\begin{subfigure}{0.5\textwidth}
\centering
\includegraphics[width=0.9\linewidth]{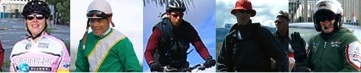}
\caption{Highest scoring results for people wearing a hat.}
\end{subfigure}
\begin{subfigure}{0.5\textwidth}
\centering
\includegraphics[width=0.9\linewidth]{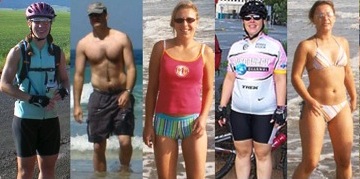}
\caption{Highest scoring results for people wearing short pants.}
\end{subfigure}
\begin{subfigure}{0.5\textwidth}
\centering
\includegraphics[width=0.9\linewidth]{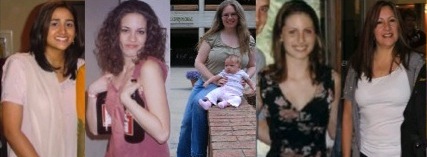}
\caption{Lowest scoring results for men.}
\end{subfigure}
\caption{Examples of highest and lowest scoring PANDA results on Berkeley Attributes of People Dataset.  The images are cropped so the required attribute can be seen better. }\label{fig:panda_results}
\end{figure}

We show some examples of highest and lowest scoring PANDA results for some attributes in Figure \ref{fig:panda_results}. Figure \ref{fig:failure} shows the top failure cases for wearing tshirts and having short hair on the test dataset. In the case of wearing tshirt, the top failure cases are picking the sleeveless, which look very similar to tshirts. And for the case of short hair, some failures are due to the person having a pony tail or the long hair is occluded in the image.

\begin{table*}[t]
\centering
\begin{tabular}{|c|c|c|c|c|c|c|c|c|c|c|}
\hline
Attribute	&	male	&long hair	&glasses	&hat 	&tshirt	& longsleeves	&short 	& jeans	& long pants & Mean AP \\
\hline
DL-Pure &  80.65 & 63.23 & 30.74 & 57.21& 37.99 & 71.76& 35.05& 60.18& 86.17 & 58.11 \\
\hline
DeCAF & 79.64 &62.29 & 31.29 & 55.17 & 41.84 & 78.77 & \textbf{80.66} & \textbf{81.46} &96.32 & 67.49 \\
\hline
Poselets150 L2 &81.70&67.07	& 44.24 & 54.01 & 42.16 & 71.70 & 36.71 & 42.56 & 87.41 & 58.62\\
\hline
DLPoselets & \textbf{92.10} & 82.26 & \textbf{76.25} & 65.55 & 44.83 & 77.31 & 43.71 & 52.52 & 87.82 & 69.15 \\
\hline
PANDA  & 91.66 & \textbf{82.70} & 69.95 & \textbf{74.22} & \textbf{49.84} & \textbf{86.01} & 79.08 & 80.99 & \textbf{96.37} & \textbf{78.98} \\
\hline
\end{tabular}
\caption{Relative performance of baselines and components of our system on the Berkeley Attributes of People test set.} \label{table-decomposed}
\end{table*}

\begin{table*}[t]
\centering
\begin{tabular}{|c|c|c|c|c|c|c|c|c|c|c|}
\hline
Partition	&	male	&long hair	&glasses	&hat 	&tshirt	& longsleeves	&shorts 	& jeans	& long pants & Mean AP \\
\hline
Frontal & 92.55 & 88.40 & 77.09 & 74.40 & 51.69 & 86.84 & 78.00 & 79.63 & 95.70 & 80.47 \\
\hline
Profile &  91.42 & 59.38 & 37.06 & 69.47 & 49.02 & 84.61 & 85.57 & 82.71 & 98.10 & 73.04 \\
\hline
Back-facing &  88.65 & 63.77 & 72.61 & 72.19 & 55.20 & 84.32 & 74.01 & 86.12 & 96.68 & 77.06 \\
\hline
All  &  91.66 & 82.70 & 69.95 & 74.22 & 49.84 & 86.01 & 79.08 & 80.99 & 96.37 &  78.98 \\
\hline
\end{tabular}
\caption{Performance of PANDA on front-facing, profile-facing and back-facing examples of the Berkeley Attributes of People test set.}
\label{table-viewpoint}
\end{table*}
\vspace{-5pt}

\subsection{Results on the Attributes25K Dataset}

Table~\ref{table-25k} shows results on the Attributes25K-test Dataset. 

\textbf{Poselets150} shows the performance of our implementation of the three-layer feed-forward network proposed by \cite{AttributesPoseletsICCV2011}. Instead of the 1200 poselets in that paper we used the 150 publicly released poselets, and instead of multiple aspect ratios we use 64x64 patches. Our system underperforms ~\cite{AttributesPoseletsICCV2011} and on the Berkeley Attributes of People dataset yields mean AP of 60.6 vs 65.2, but it is faster and simpler and we have adopted the same setup for our CNN-based poselets. This allows us to make more meaningful comparisons between the two methods.

\textbf{DPD} and \textbf{DeCAF}  We used the publicly available implementations of~\cite{AttributesDPMsICCV2013} based on deformable part models and~\cite{decaf} based on CNN trained on ImageNet.

\textbf{DL-DPM} shows the performance of using DPM parts instead of poselets. We used the pretrained DPM model in~\cite{AttributesDPMsICCV2013}. We extracted the patches associated with each of the 8 parts in each of the 6 mixture components for a total of 48 parts (only 8 of which are active at a time). We then used the same setup as PANDA -- trained CNN classifiers for each of the 48 parts, combined them with the global model and trained SVM on top. As the table shows, \textbf{DL-DPM} outperforms conventional part-based methods (both DPM and poselets) which do not use deep features as well as \textbf{DeCAF}. However it does not match the performance of the deep poselets in PANDA. The patches from DPM parts have higher noise (because they have to fire even if the pattern is weak or non-existent) and are not well aligned (because they have to satisfy global location constraints).

\subsection{Component Evaluation}

We now explore the performance of individual components of our system as shown on Table~\ref{table-decomposed} using the Berkeley dataset. Our goal is to get insights into the importance of using deep learning and the importance of using parts.

\emph{How well does a conventional deep learning classifier perform?} We first explore a simple model of feeding the raw RGB image of the person into a deep network. To help with rough alignment and get signal from two resolutions we split the images into four 64x64 patches -- one from the top, center, and bottom part of the person's bounds, and one from the full bounding box at half the resolution. In total we have 4 concatenated 64x64 square color images as input (12 channels).
We train a CNN on this 12x64x64 input on the full Attributes-25K dataset.  The structure we used is similar to the CNN in Figure \ref{fig:outline_hybrid} and it consists of 
two convolution/normalization/pooling stages,
 followed by a fully connected layer with 512 hidden units followed by nine columns, each composed of one hidden layer with 128 hidden units. Each of the 9 branches outputs a single value which is a binary classifier of the attribute.
We then use the CNN as a feature extractor on the validation set by using the features produced by the final fully connected layer. We train a logistic regression using these features and report its performance on the ICCV test set as ~\textbf{DL-Pure} on Table~\ref{table-decomposed}. 

We also show the results of our second baseline -- DeCAF, which is the global component of our system. Even though it is a convolutional neural net originally trained on a completely different problem (ImageNet classification), it has been exposed to millions of images and it outperforms~\textbf{DL-Pure}.

\emph{How important is deep learning at the part level?}. By comparing the results of~\textbf{Poselets150L2} and~\textbf{DLPoselets} we can see the effect of deep learning at the part level. Both methods use the same poselets, train poselet-level attribute classifiers and combine them at the person level with a linear SVM. The only difference is that Poselets150L2 uses the features as described in ~\cite{AttributesPoseletsICCV2011} (HOG features, color histogram, skin tone and part masks) whereas DLPoselets uses features trained with a convolutional neural net applied to the poselet image patch. As our table shows, deep-net poselets result in increased performance.

~\textbf{PANDA} shows the results of our proposed system which combines DeCAF and DLPoselets. Our part and holistic classifiers use complementary features and combining them together further boosts the performance.

\subsection{Robustness to viewpoint variations}

In Table~\ref{table-viewpoint}, we show the performance of our method as a function of the viewpoint of the person. We considered as \emph{frontal} any image in which both eyes of the person are visible, which includes approximately 60\% of the dataset. \emph{Profile} views are views in which one eye is visible and \emph{Back-facing} are views where both eyes are not visible.  As expected, our method performs best for front-facing people because they are most frequent in our training set. However, the figure shows that PANDA can work well across a wide range of viewpoints.

%
%
%

\subsection{Results on the LFW Dataset}
We also report results on the Labeled Faces in the Wild dataset~\cite{lfw_iccv09}. The dataset consists of 13233 images of cropped, centered frontal faces. The registered nature of the data does not leverage the strengths of our system in its ability to deal with viewpoint, pose and partial occlusions. Nevertheless, it provides us another datapoint to compare against other methods. This dataset contains many attributes, but unfortunately the ground truth labels are not released. We used crowd-sourcing to collect ground-truth labels for the gender attribute only. We split the examples randomly into 3042 training and 10101 test examples with the only constraint that the same identity may not appear in both training and test sets. We used our system whose features were trained on Attribute-25K to extract features on the 3042 training examples. Then we trained a linear SVM and applied the classifier on the 10101 test examples. We also used the publicly available gender scores of~\cite{lfw_iccv09} to compute the average precision of their system on the test subset. The results are shown on Table~\ref{table-lfw}.

\begin{table}[t]
\centering
\begin{tabular}{|c|c|}
\hline
Method &  Gender AP  \\
\hline
 Simile~\cite{lfw_iccv09} & 95.52 \\
\hline
FrontalFace poselet & 96.43 \\
\hline
PANDA & \textbf{99.54} \\
\hline
\end{tabular}
\caption{Average precision of PANDA on the gender recognition of the LFW dataset.}
\label{table-lfw}
\end{table}
\begin{figure}[t]
\centering
\includegraphics[width=0.9\linewidth]{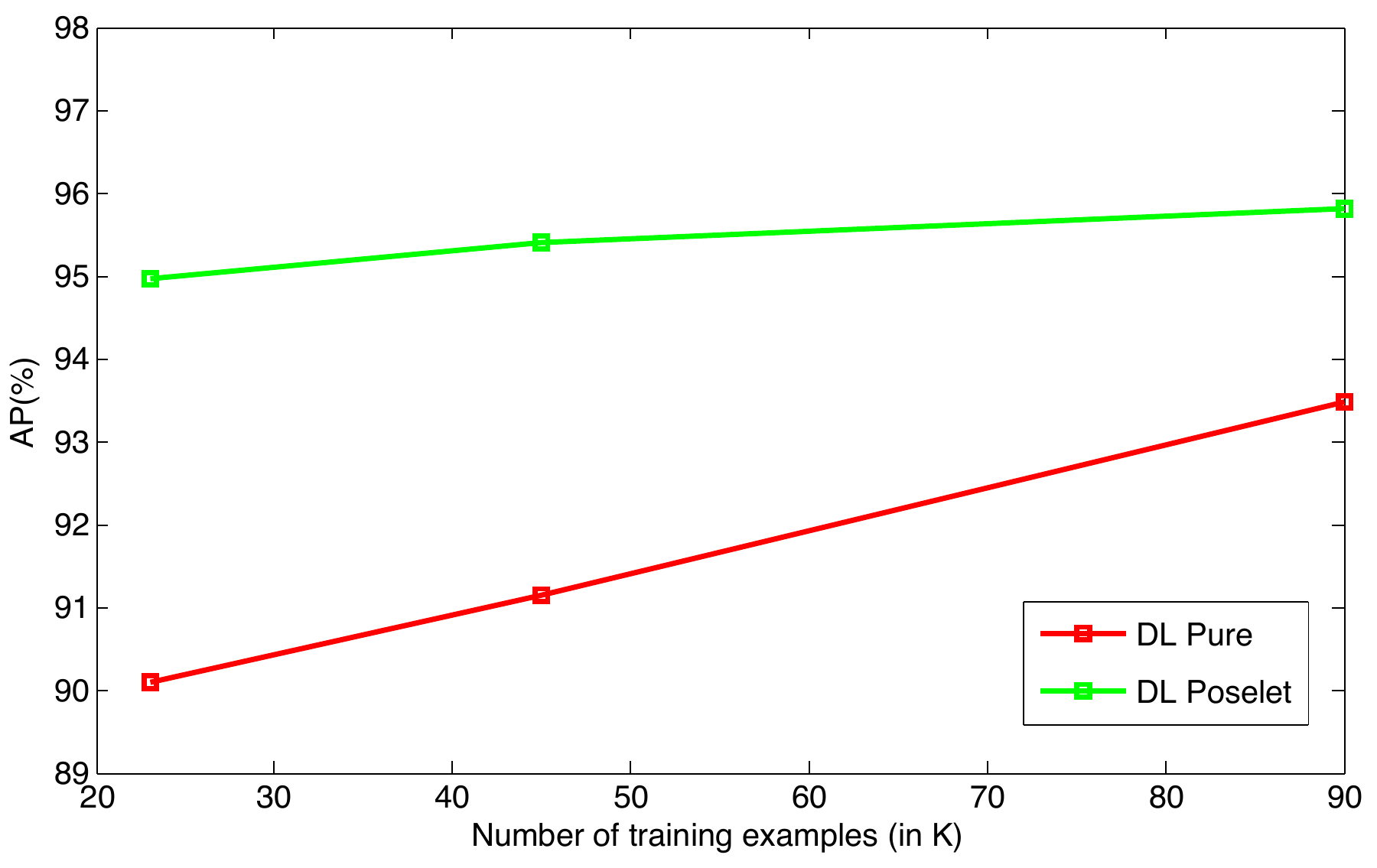}
\caption{Average precision of gender classification as a function of the training size. }
 \label{fig:large_scale}
\end{figure}

PANDA's AP on LFW is 99.54\% using our parts model, a marked improvement over the previous state of the art. Our manual examination of the results shows that roughly 1 in 200 test examples either had the wrong ground truth or we failed to match the detection results with the correct person. Thus PANDA achieves nearly perfect gender recognition performance in LFW and it shows that PANDA is effective even when images are tightly cropped and variation in pose is reduced. 

One interesting observation is that, even though the dataset consists of only frontal-face people, the performance of our frontal-face poselet is significantly lower than the performance of the full system. This suggests that our system benefits from combining the signal from multiple  redundant  classifiers, each of which is trained on slightly different set of images.

\subsection{Analysis of dataset size}
We also investigate the effects of number of training examples for DL Pure and DL Poselet methods. It is interesting to see if holistic deep learning method trained on the whole bounding box image can deal with pose variations implicitly given a large amount of data. 
We collected a dataset for gender classification, consisting of \tilde90K  training, \tilde2.6K validation and \tilde10K  test examples of people from photo albums. The ground truth labels have about 1.5\% noise. We trained on the full training set of 90K, and on subsets of 45K and 23.5K. The number of training examples for PANDA (poselet activations) are 5.6 million, 2.8 million and 1.4 million respectively. We followed the same pipeline by using the same set of poselets and same part level convolutional neural nets in the experiments above. The results are shown in Figure~\ref{fig:large_scale}. The holistic model (DL Pure) has an almost linear improvement over the number of training examples while our pose aligned method outperforms the holistic method but having a smaller improvement as the training size increases. We would like to experiment on a larger dataset to see if those two methods can intersect in the future.

\section{Conclusion}

We presented a method for attribute classification of people that improves performance compared with previously published methods. It is conceptually simple and leverages the strength of convolutional neural nets without requiring datasets of millions of images. It uses poselets to factor out the pose and viewpoint variation which allows the convolutional network to focus on the pose-normalized appearance differences. We concatenate the deep features at each poselet and add a deep representation of the whole input image. Our feature representation is generic and we achieve state-of-the-art results on the Berkeley Attributes of People dataset and on LFW even if we train our CNNs on a different dataset. We believe that our proposed hybrid method using mid-level parts and deep learning classifiers at each part will prove effective not just for attribute classification, but also for problems such as detection, pose estimation, action recognition.

\paragraph{Acknowledgments} The first and the fourth authors were supported in part by DARPA Mind's Eye and MSEE
programs, by NSF awards IIS-0905647, IIS-1134072, and IIS-1212798, and
by support from Toyota.

\balance
{\small
\bibliographystyle{ieee}
\bibliography{attrbib}
}

\end{document}